# Brain Predictability toolbox: a Python library for neuroimaging based machine learning

Hahn, S.[1], Yuan, D.K.[1], Thompson, W.K.[2], Owens, M.[1], Allgaier, N[1]. and Garavan, H[1]

1. Departments of Psychiatry and Complex Systems, University of Vermont, Burlington, VT 05401, USA
2. Division of Biostatistics, Department of Family Medicine and Public Health, University of California, San Diego, La Jolla, CA 92093, USA

# Abstract

**Summary**
Brain Predictability toolbox (BPt) represents a unified framework of machine learning (ML) tools designed to work with both tabulated data (in particular brain, psychiatric, behavioral, and physiological variables) and neuroimaging specific derived data (e.g., brain volumes and surfaces). This package is suitable for investigating a wide range of different neuroimaging based ML questions, in particular, those queried from large human datasets.

**Availability and Implementation**
BPt has been developed as an open-source Python 3.6+ package hosted at https://github.com/sahahn/BPt under MIT License, with documentation provided at https://bpt.readthedocs.io/en/latest/, and continues to be actively developed. The project can be downloaded through the github link provided. A web GUI interface based on the same code is currently under development and can be set up through docker with instructions at https://github.com/sahahn/BPt_app.

**Contact**
Please contact Sage Hahn at sahahn@uvm.edu

# Main Text

## 1 Introduction

Large datasets in all domains are becoming increasingly prevalent as data from smaller existing studies are pooled and larger studies are funded. This increase in available data offers an unprecedented opportunity for researchers interested in applying machine learning (ML) based methodologies, especially those working in domains such as neuroimaging where data collection is quite expensive. This paper considers neuroimaging based ML (analyses of brain data) as an example domain in which the toolbox can be applied.

While there are a number of existing libraries for performing general ML based workflows within Python and other languages, the Brain Predictability toolbox (BPt) offers a high level user interface with specific consideration made towards neuroimaging based ML. BPt is designed to supplement the experience currently offered by similar popular libraries such as scikit-learn (Pedregosa, 2011) and nilearn (Abraham, 2014), rather than replace. BPt leverages existing ML libraries along with new functionality in order to provide a resource suitable for guiding users through the full research ML workflow; from loading data to interpreting results.

## 2 Description

### 2.1 Usability
BPt offers both a python based api and a web interface application, each with overlapping utility and distinct strengths and weaknesses. In this way, BPt seeks to balance "user friendliness" and expressiveness, with the goal of creating a framework suitable for both beginners, and one with enough flexibility to be used by advanced ML practitioners. That said, this library is not explicitly designed as a tutorial for new users. Some baseline knowledge of machine learning is required as well as some background Python knowledge, though the web interface version of the project (BPt_app) seeks to eliminate the latter prerequisite. A comprehensive documentation is provided along with a number of detailed examples for the Python api. Examples, found at https://github.com/sahahn/BPt/tree/master/Examples, are provided as jupyter notebooks and explore a range of problem types on real world data.

### 2.2 Best Practices
The underlying structure of the library guides users to follow best practices in regard to cross validation, namely; perform a global train-test split, using the training set for model pipeline exploration and ultimately evaluating on the testing set. Performance from each step is easily reported over multiple user-defined metrics. The general structure of both the library and web application further guides users through a recommended workflow.

### 2.3 Data Loading

BPt allows a user to easily load, manipulate and interactively view input neuroimaging datasets. Loading functions are equipped to help perform outlier detection, handling of missing data, loading of specific variables and detection of duplicate variables among a number of other utilities. Data visualization tools are implemented in order to facilitate active data exploration.

**2.4 ML Pipelines**
Diverse and complex ML pipelines can easily be created with a number of predefined choices across a range of state of the art ML techniques. BPt strives to include as broad and as recent a selection of different ML algorithms as possible, as well as to directly integrate these choices with custom and preset hyperparameter distributions. Users can further express the choice between one or more algorithms or pipeline steps as hyperparams, allowing for the easy inclusion of model selection as properly nested within cross-validation.

**2.5 Problem Type Support**
All common ML problem types are supported (regression, binary and categorical), with low level implementation issues abstracted away, and new wrapper functions written to provide extended problem type support.

**2.6 Covariates and Feature Importance**
Properly handling covariates within neuroimaging based machine learning is rarely straightforward. BPt supports a range of techniques for estimating the influence of covariates, including: feature importance, leave-out group CV (e.g., leave-out site for multi-site neuroimaging data), experiments on one group (e.g., sex-specific classifier), post-stratifying raw predictions (e.g., by race) and others. Feature importance in particular is supported by extracting base measures (e.g., beta weights from linear models), in addition to calculating SHapley Additive exPlanations (Lundberg, 2017) and permutation-derived feature importances (Altmann, 2010).

**2.7 Reproducibility**
By conducting loading, preprocessing and modeling within the same script, analyses can be easily reproduced and shared. Automatic logs are generated within the python workflow and similarly within the web app projects can be easily created and saved. These tools allow previous analyses to be easily retrievable.

**2.8 Convenience**
Most researchers working on neuroimaging based ML applications, or other applied academic ML, have little background in software engineering, which means that writing code for loading data and building ML models can often take longer than expected or introduce unexpected bugs. Instead, by leveraging BPt, researchers can quickly move from ideas to experimentation and, importantly, results.

**2.9 Backend Libraries**

BPt makes use of a number of other libraries within the scientific Python community, which without their contribution this project would not be possible, most notably: Numpy (Oliphant, 2006), pandas (McKinney, 2010) and scikit-learn (Pedregosa, 2011). Plotting functionality makes use of the matplotlib library (Hunter, 2007). Extra classifiers and pipeline objects beyond those included with scikit-learn are used from python libraries: lightgbm (Ke, 2017), xgboost (Chen, 2016), imbalanced-learn (Lemaître, 2017), and DESlib (Cruz 2018). Hyperparameter optimizers are implemented through FaceBook's Nevergrad library, and additional feature importance support is added with the Shap library (Lundberg, 2017).

## Acknowledgements
We would like to thank the members of the Hugh Garavan lab for their assistance in testing the library. We would further like to thank the Data Analysis and Informatic Core of the ABCD Study who provided the structure of the code in which the web interface application was developed from.

## Funding information
This work was funded in part by NIDA grant T32DA043593.  All authors report no conflicts of interest.